\documentclass{article}

\usepackage{arxiv}

\usepackage[utf8]{inputenc} 
\usepackage[T1]{fontenc}    
\usepackage{hyperref}       
\usepackage{url}            
\usepackage{booktabs}       
\usepackage{amsfonts}       
\usepackage{nicefrac}       
\usepackage{microtype}      
\usepackage{lipsum}
\usepackage{fancyhdr}       
\usepackage{graphicx}       
\graphicspath{{media/}}     
\usepackage{amsmath}    
\usepackage{algorithm}  
\usepackage{algpseudocode} 
\usepackage{pgfplots}  
\usepackage{tikz}      
\usepackage{amsmath}   
\usepackage{booktabs}  
\usepackage{graphicx}  
\usepackage{float}     
\usepackage[numbers]{natbib}

\graphicspath{ {./images/} }

\title{Entropy Adaptive Decoding: Dynamic Model Switching for Efficient Inference 

}

\author{
Toby Simonds \\\\
Tufa Labs \\\\
\texttt{toby@tufalabs.ai}
}

\begin{document}
\maketitle

\begin{abstract}
We present Entropy Adaptive Decoding (EAD), a novel approach for efficient language model inference that dynamically switches between different-sized models based on prediction uncertainty. By monitoring rolling entropy in model logit distributions, our method identifies text regions where a smaller model suffices and switches to a larger model only when prediction uncertainty exceeds a threshold. Unlike speculative decoding approaches that maintain perfect output fidelity through verification, EAD accepts controlled output divergence in exchange for computational efficiency. Our experiments on the MATH benchmark demonstrate remarkable efficiency gains across different model families. Using the LLaMA family, we maintain 96.7\% of the 11B model's performance (50.4\% vs 52.1\%) while using it for only 43\% of tokens, decreasing computational cost by 41.5\%. These gains become more pronounced with larger size differentials in the Qwen family, where we achieve 92.9\% of the 14B model's performance (74.3\% vs 80.0\%) while using it for just 25\% of tokens, decreasing computational cost by 67\%. The consistency of these results across model pairs suggests that language model computation can be significantly optimized by selectively deploying model capacity based on local generation complexity. Our findings indicate that current approaches to model inference may be unnecessarily conservative in their pursuit of perfect output fidelity, and that accepting minor performance trade-offs can enable dramatic reductions in computational costs.\end{abstract}

\section{Introduction}
Large language models (LLMs) have demonstrated unprecedented capabilities in diverse cognitive tasks, yet their substantial computational requirements during inference pose a significant implementation challenge. While recent optimization techniques, particularly speculative decoding, have achieved notable improvements in inference latency through parallel token prediction, these methods maintain strict output equivalence by verifying all predictions against a reference model. This conservative approach to maintaining fidelity potentially foregoes opportunities for additional computational optimization.

A fundamental observation in language model behavior reveals that text generation exhibits variable complexity requirements. Routine linguistic patterns and predictable continuations can be adequately handled by models of modest capacity, while complex reasoning tasks and domain-specific applications may necessitate more substantial computational resources. This heterogeneity in computational requirements suggests an opportunity for dynamic resource allocation based on the local complexity of the generation task. In this context, entropy emerges as a computationally efficient metric for approximating generation difficulty - elevated entropy in the model's prediction distribution indicates uncertainty that may benefit from increased model capacity, while reduced entropy suggests simpler, more predictable generations amenable to smaller models.

This inefficiency becomes particularly pronounced in modern Chain-of-Thought (CoT) models\cite{noauthor_openai_nodate} that generate thousands of intermediate reasoning tokens. When these models tackle complex tasks, they often produce extensive step-by-step explanations where a small fraction of tokens represent crucial logical deductions while the majority consist of explanatory scaffolding. For example, in a mathematical proof, the key breakthrough might lie in recognizing a specific substitution or transformation, yet the model expends equal computation on generating routine explanatory phrases like "this means that" or "we can conclude." This misallocation of computational resources scales linearly with the length of the reasoning chain, resulting in significant inefficiencies as models generate increasingly verbose step-by-step explanations. The current paradigm of equal computation per token thus becomes particularly problematic in extended CoT models, where thousands of tokens may be generated but only a small subset truly drives the reasoning process forward.

We present Entropy Adaptive Decoding (EAD), a novel approach for dynamic model switching during inference based on prediction uncertainty. Rather than pursuing perfect output fidelity through verification mechanisms as in speculative decoding, EAD explicitly trades bounded output divergence for computational efficiency. The system monitors model uncertainty through a rolling window of entropy values computed from prediction logits, enabling it to identify text regions where a smaller model's capacity suffices and switching to a larger model only when uncertainty exceeds defined thresholds.

This methodology represents a fundamental shift from current optimization paradigms. Instead of using smaller models for speculative prediction with subsequent verification, EAD directly generates from models of different capacities based on local entropy metrics. This allows for controlled divergence from the larger model's behavior in exchange for substantial computational gains. The dynamic switching mechanism continuously adapts to the local complexity of the generation task, ensuring computational resources are allocated proportionally to the predicted difficulty of each text region.

Through extensive empirical evaluation, we demonstrate that EAD achieves significant computational efficiency while maintaining performance within acceptable bounds. Our results on the MATH benchmark suggest that the strict output equivalence enforced by current speculative decoding approaches may be unnecessarily conservative. By accepting minor deviations in output, we can achieve dramatic reductions in computational cost while preserving the majority of larger model performance benefits. These findings challenge the assumption that perfect output fidelity is necessary for practical applications and suggest a new direction for efficient language model deployment.

\section{Related Work}
\subsection{Uncertainty Estimation in Language Models}
Uncertainty estimation in neural language models has been extensively studied through various lens, with entropy in logit distributions emerging as a key metric. Early work by \cite{shannon_prediction_1951} established the connection between entropy and predictability in natural language, while more recent studies have applied these insights to neural models. The use of entropy as an uncertainty metric has found applications beyond pure measurement. \cite{holtzman_curious_2020} showed that locally dynamic temperature sampling based on entropy can improve generation quality. \cite{meister_locally_2023} used entropy in beam search to better balance exploration and exploitation during decoding. These works establish entropy as a reliable proxy for generation difficulty, though they primarily focus on improving quality rather than computational efficiency.
\subsection{Speculative Decoding}
Speculative decoding has emerged as a promising approach for accelerating language model inference. The technique was first introduced by \cite{leviathan_fast_2023}. Their key insight was that smaller helper models could predict multiple tokens in parallel, with verification against a larger target model ensuring output quality. The Medusa architecture \citep{cai_medusa_2024} attached specialized prediction heads to existing models rather than using separate helper models. These approaches all maintain perfect output equivalence with the target model through verification mechanisms, prioritizing quality over potential computational savings.
\subsection{Adaptive Computation in Language Models}
The concept of dynamically adjusting computational resources has precedent in language model research. \cite{chen_ee-llm_2024} proposed early-exit transformers that could terminate computation at different layers based on confidence thresholds. These approaches typically focus on layer-wise adaptivity within a single model rather than switching between models of different sizes. There has also been work on switching between different smaller models that are tuned to specific tasks to save compute \cite{simonds_modem_2024}'s. Our work bridges these research areas by using entropy-based uncertainty estimation to guide adaptive model switching during generation. Unlike previous approaches that maintain perfect output equivalence or adapt computation within a single model, we explore the novel direction of allowing controlled output divergence in exchange for computational benefits.

\section{Methodology}

\subsection{Entropy-Based Model Selection}
Our system consists of two language models of different parameter counts: a smaller model $M_S$ and a larger model $M_L$. The models can be from the same family (e.g., different sizes of LLaMA) or different architectures entirely.

The core of our approach lies in using entropy as a proxy for prediction difficulty. For a given sequence of tokens $x_{1:t}$, each model produces logits $\mathbf{l} \in \mathbb{R}^V$ representing the unnormalized probabilities for the next token. We calculate entropy before any temperature scaling to capture the model's true uncertainty:

\begin{align}
    p_i &= \frac{\exp(l_i)}{\sum_{j=1}^V \exp(l_j)} \\
    H_t &= -\sum_{i=1}^V p_i \log_2(p_i)
\end{align}

To smooth local fluctuations, we maintain a rolling window of size $w$ and compute the average entropy:

\begin{equation}
    \bar{H}_t = \frac{1}{w}\sum_{i=t-w+1}^t H_i
\end{equation}

\subsection{Dynamic Switching Mechanism}
Model selection at each timestep $t$ is determined by comparing the average entropy to a threshold $\tau$:

\begin{equation}
    M_t = \begin{cases}
        M_L & \text{if } \bar{H}_t > \tau \text{ and } c \geq d_{min} \\
        M_S & \text{if } \bar{H}_t \leq \tau \text{ and } c \geq d_{min} \\
        M_{t-1} & \text{otherwise}
    \end{cases}
\end{equation}

where $c$ counts tokens generated by the current model and $d_{min}$ is the minimum number of tokens required before allowing a switch. This prevents rapid oscillation between models and ensures stable generation.

\begin{algorithm}
\caption{Adaptive Speculative Decoding Generation Process}
\begin{algorithmic}[1]
\Require Initial prompt sequence $x$, models $M_S$, $M_L$, entropy threshold $\tau$, window size $w$, minimum switch duration $d_{min}$, temperature $T$
\Ensure Generated sequence $y$
\State Initialize empty token sequence $y \gets []$
\State Initialize entropy window $W \gets []$
\State Initialize token counter $c \gets 0$
\State Initialize current model $M_t \gets M_S$
\While{not done}
    \State $\mathbf{l} \gets M_t(x + y)$ \Comment{Generate logits}
    \State $p \gets \text{softmax}(\mathbf{l})$
    \State $H_t \gets -\sum_{i} p_i \log_2(p_i)$ \Comment{Calculate entropy}
    \State Append $H_t$ to $W$
    \State $\bar{H_t} \gets \text{mean}(W[-w:])$ \Comment{Update rolling average}
    \State $\mathbf{l'} \gets \mathbf{l}/T$ \Comment{Apply temperature}
    \State $p' \gets \text{softmax}(\mathbf{l'})$
    \State $t \gets \text{sample}(p')$ \Comment{Sample next token}
    \State Append $t$ to $y$
    \State $c \gets c + 1$
    \If{$c \geq d_{min}$}
        \If{$M_t = M_S$ and $\bar{H_t} > \tau$}
            \State $M_t \gets M_L$
            \State $c \gets 0$
        \ElsIf{$M_t = M_L$ and $\bar{H_t} \leq \tau$}
            \State $M_t \gets M_S$
            \State $c \gets 0$
        \EndIf
    \EndIf
\EndWhile
\Return $y$
\end{algorithmic}
\end{algorithm}

\subsection{Implementation Details}
Our implementation includes several practical considerations:
\paragraph{Hyperparameters} Key parameters include:
\begin{itemize}
    \item Entropy threshold $\tau$: Controls switching sensitivity
    \item Window size $w$: For entropy smoothing
    \item Minimum switch duration $d_{min}$: Prevents oscillation
\end{itemize}

In our experiments, we used a window size $w=5$ for entropy smoothing and a minimum switch duration $d_{min}=10$ tokens to prevent rapid oscillation between models. We explore various entropy thresholds $\tau$ in our results section to understand the trade-off between computational efficiency and output quality. While these parameters proved effective in our experiments, further work could optimize these hyperparameters for specific applications - for instance, creative writing might benefit from different switching thresholds compared to mathematical reasoning tasks.

 We evaluate our approach on the MATH benchmark \citep{hendrycks_measuring_2021}, a dataset of 12,500 problems which tests advanced mathematical reasoning capabilities. We evaluate our approach on a 1000 problem subset fo the MATH benchmark.

\section{Results}

We tested various entropy thresholds ($\tau$) to understand the trade-off between computational efficiency and performance. Each threshold results in a different ratio of small-to-large model usage, effectively creating a spectrum of computational costs.

\begin{figure}[!htbp]
\centering
\begin{tikzpicture}
\begin{axis}[name=plot1,
   at={(-2cm,5cm)},
   width=8cm, height=8cm,
   xlabel={Large Model Usage (\%)},
   ylabel={MATH Score (\%)},
   xmin=0, xmax=100, ymin=25, ymax=50,
   title={LLama 3.2 1B vs LLama 3.2 3B},
   grid=major,
   grid style={gray!20},
   axis lines=left,
   tick style={color=black}
]
\addplot[only marks, mark=*, mark size=3pt, color={rgb:red,0.7;blue,0}] coordinates {(100.0,46.4)};
\addplot[only marks, mark=*, mark size=3pt, color={rgb:red,0.6;blue,0.2}] coordinates {(57.0,44.5)};
\addplot[only marks, mark=*, mark size=3pt, color={rgb:red,0.5;blue,0.3}] coordinates {(45.0,45.2)};
\addplot[only marks, mark=*, mark size=3pt, color={rgb:red,0.4;blue,0.4}] coordinates {(38.0,41.7)};
\addplot[only marks, mark=*, mark size=3pt, color={rgb:red,0.3;blue,0.5}] coordinates {(32.0,41.0)};
\addplot[only marks, mark=*, mark size=3pt, color={rgb:red,0.2;blue,0.6}] coordinates {(23.3,41.9)};
\addplot[only marks, mark=*, mark size=3pt, color={rgb:red,0.1;blue,0.7}] coordinates {(16.0,36.0)};
\addplot[only marks, mark=*, mark size=3pt, color={rgb:red,0;blue,0.8}] coordinates {(0.0,30.4)};

\end{axis}

\begin{axis}[name=plot2,
   at={(7cm,5cm)},
   width=8cm, height=8cm,
   xlabel={Large Model Usage (\%)},
   ylabel={MATH Score (\%)},
   xmin=0, xmax=100, ymin=46, ymax=54,
   title={LLama 3.2 3B vs LLama 3.2 11B},
   grid=major,
   grid style={gray!20},
   axis lines=left,
   tick style={color=black}
]
\addplot[only marks, mark=*, mark size=3pt, color={rgb:red,0.7;blue,0}] coordinates {(100.0,52.0)};
\addplot[only marks, mark=*, mark size=3pt, color={rgb:red,0.6;blue,0.2}] coordinates {(58.0,51.4)};
\addplot[only marks, mark=*, mark size=3pt, color={rgb:red,0.5;blue,0.3}] coordinates {(46.0,51.1)};
\addplot[only marks, mark=*, mark size=3pt, color={rgb:red,0.4;blue,0.4}] coordinates {(43.0,50.4)};
\addplot[only marks, mark=*, mark size=3pt, color={rgb:red,0.3;blue,0.5}] coordinates {(33.0,50.3)};
\addplot[only marks, mark=*, mark size=3pt, color={rgb:red,0.2;blue,0.6}] coordinates {(26.0,49.0)};
\addplot[only marks, mark=*, mark size=3pt, color={rgb:red,0.1;blue,0.7}] coordinates {(16.0,48.4)};
\addplot[only marks, mark=*, mark size=3pt, color={rgb:red,0;blue,0.8}] coordinates {(0.0,48.0)};

\end{axis}

\begin{axis}[name=plot3,
   at={(-2cm,-4cm)},
   width=8cm, height=8cm,
   xlabel={Large Model Usage (\%)},
   ylabel={MATH Score (\%)},
   xmin=0, xmax=100, ymin=25, ymax=80,
   title={Qwen 2.5 0.5B vs Qwen 2.5 14B},
   grid=major,
   grid style={gray!20},
   axis lines=left,
   tick style={color=black}
]
\addplot[only marks, mark=*, mark size=3pt, color={rgb:red,0;blue,0.8}] coordinates {(0.0,34.4)};
\addplot[only marks, mark=*, mark size=3pt, color={rgb:red,0.1;blue,0.7}] coordinates {(8.3,46.0)};
\addplot[only marks, mark=*, mark size=3pt, color={rgb:red,0.2;blue,0.6}] coordinates {(14.9,59.5)};
\addplot[only marks, mark=*, mark size=3pt, color={rgb:red,0.3;blue,0.5}] coordinates {(22.1,63.2)};
\addplot[only marks, mark=*, mark size=3pt, color={rgb:red,0.4;blue,0.4}] coordinates {(27.2,64.1)};
\addplot[only marks, mark=*, mark size=3pt, color={rgb:red,0.5;blue,0.3}] coordinates {(35.3,69.0)};
\addplot[only marks, mark=*, mark size=3pt, color={rgb:red,0.6;blue,0.2}] coordinates {(43.2,70.0)};
\addplot[only marks, mark=*, mark size=3pt, color={rgb:red,0.7;blue,0}] coordinates {(100.0,80.0)};

\end{axis}

\begin{axis}[name=plot4,
   at={(7cm,-4cm)},
   width=8cm, height=8cm,
   xlabel={Large Model Usage (\%)},
   ylabel={MATH Score (\%)},
   xmin=0, xmax=100, ymin=30, ymax=85,
   title={Qwen 2.5 1.5B vs Qwen 2.5 14B},
   grid=major,
   grid style={gray!20},
   axis lines=left,
   tick style={color=black}
]
\addplot[only marks, mark=*, mark size=3pt, color={rgb:red,0;blue,0.8}] coordinates {(0,34.4)};
\addplot[only marks, mark=*, mark size=3pt, color={rgb:red,0.2;blue,0.6}] coordinates {(9.9,67.5)};
\addplot[only marks, mark=*, mark size=3pt, color={rgb:red,0.3;blue,0.5}] coordinates {(18.5,71.7)};
\addplot[only marks, mark=*, mark size=3pt, color={rgb:red,0.4;blue,0.4}] coordinates {(25.0,74.3)};
\addplot[only marks, mark=*, mark size=3pt, color={rgb:red,0.5;blue,0.3}] coordinates {(31.0,73.2)};
\addplot[only marks, mark=*, mark size=3pt, color={rgb:red,0.6;blue,0.2}] coordinates {(33.0,74.9)};
\addplot[only marks, mark=*, mark size=3pt, color={rgb:red,0.7;blue,0.1}] coordinates {(40.0,75.5)};
\addplot[only marks, mark=*, mark size=3pt, color={rgb:red,0.8;blue,0}] coordinates {(100.0,80.0)};
\end{axis}

\node[below] at (5cm,-6cm) {
   \begin{tabular}{cccccccc}
       \textcolor{rgb,1:red,0.7;blue,0}{$\bullet$} $\tau=0$ &
       \textcolor{rgb,1:red,0.6;blue,0.2}{$\bullet$} $\tau=0.03125$ &
       \textcolor{rgb,1:red,0.5;blue,0.3}{$\bullet$} $\tau=0.0625$ &
       \textcolor{rgb,1:red,0.4;blue,0.4}{$\bullet$} $\tau=0.125$ &
       \textcolor{rgb,1:red,0.3;blue,0.5}{$\bullet$} $\tau=0.25$ &
       \textcolor{rgb,1:red,0.2;blue,0.6}{$\bullet$} $\tau=0.5$ &
       \textcolor{rgb,1:red,0.1;blue,0.7}{$\bullet$} $\tau=1$ &
       \textcolor{rgb,1:red,0;blue,0.8}{$\bullet$} $\tau=99$
   \end{tabular}
};
\end{tikzpicture}
\caption{Comparison between large model utilization by different entropy threshold and MATH score. Color represents entropy threshold. All models are instruction tuned variants }
\label{fig:config_grid}
\end{figure}
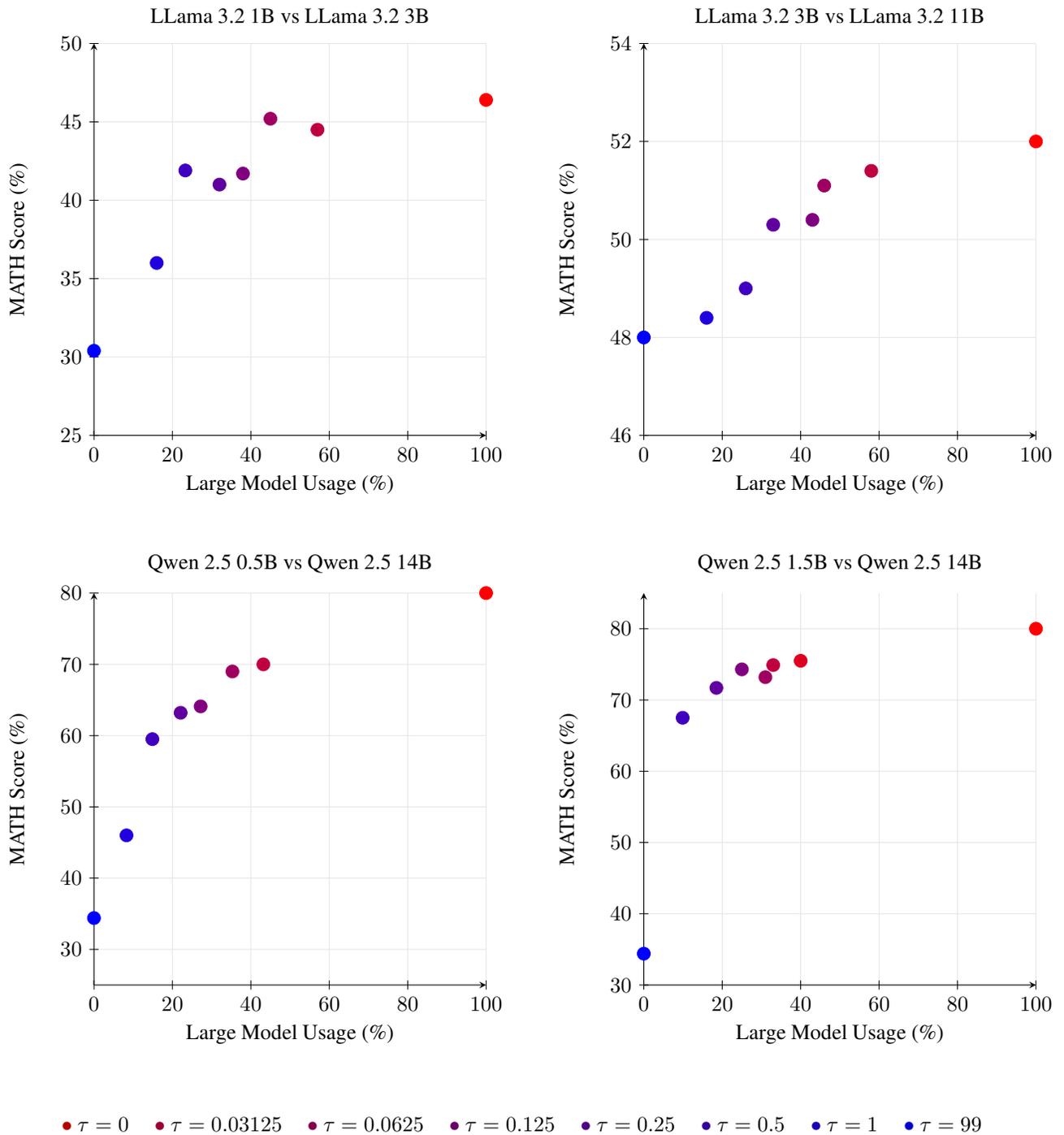

\begin{table}[ht]
\centering
\caption{Performance and Parameter Usage Across Model Pairs and Entropy Thresholds}
\label{tab:model_comparisons}
\begin{tabular}{@{}lrrr@{}}
\toprule
Entropy ($\tau$) & Large Model Usage (\%) & Score (\%) & Parameter Ratio (\%)\textsuperscript{*} \\
\midrule
\multicolumn{4}{l}{\textbf{Llama 3.2 3B vs 1B Model}} \\
\midrule
99 & 0.0 & 30.4 & 33.3 \\
1 & 16.0 & 36.0 & 44.0 \\
0.5 & 23.3 & 41.9 & 48.9 \\
0.25 & 32.0 & 41.0 & 54.7 \\
0.125 & 38.0 & 41.7 & 58.7 \\
0.0625 & 45.0 & 45.2 & 63.5 \\
0.03125 & 57.0 & 44.5 & 71.3 \\
0 & 100.0 & 46.4 & 100 \\
\midrule
\multicolumn{4}{l}{\textbf{Llama 3.2 11B vs 3B Model}} \\
\midrule
99 & 0.0 & 48.0 & 27.3 \\
1 & 16.1 & 48.4 & 38.5 \\
0.5 & 26.6 & 49.0 & 46.3 \\
0.25 & 33.3 & 50.3 & 51.4 \\
0.125 & 43.2 & 50.4 & 58.5 \\
0.0625 & 46.1 & 51.1 & 61.0 \\
0.03125 & 58.7 & 51.4 & 69.2 \\
0 & 100 & 52.0 & 100.0 \\

\midrule
\multicolumn{4}{l}{\textbf{Qwen 2.5 14B vs 0.5B Model}} \\
\midrule
99 & 0.0 & 34.4 & 3.6 \\
1 & 8.3 & 46.0 & 11.6 \\
0.5 & 14.9 & 59.5 & 17.9 \\
0.25 & 22.1 & 63.2 & 24.9 \\
0.125 & 27.2 & 64.1 & 29.8 \\
0.0625 & 35.3 & 69.0 & 37.6 \\
0.03125 & 43.2 & 70.0 & 45.2 \\
0 & 100.0 & 80.0 & 100.0 \\
\midrule
\multicolumn{4}{l}{\textbf{Qwen 2.5 14B vs 1.5B Model}} \\
\midrule
99 & 0 & 34.4 & 10.7 \\
1 & 9.9 & 67.5 & 19.6 \\
0.5 & 18.5 & 71.7 & 27.2 \\
0.25 & 25.6 & 74.3 & 33.6\\
0.125 & 31.0 & 73.2 & 38.4 \\
0.0625 & 33.3 & 74.9 & 40.4 \\
0.03125 & 40.7 & 75.5 & 47.1\\
0 & 100.0 & 80.1 & 100.0 \\
\bottomrule
\end{tabular}

\vspace{1mm}
\raggedright
\footnotesize
\textsuperscript{*}Parameter ratio represents the average number of parameters used in the forward pass compared to using only the larger model, calculated as $\alpha P_S + \beta P_L$ where $\alpha$ and $\beta$ are the usage proportions of small and large models, normalized by $P_L$.
\end{table}

Our experimental evaluation measured three key metrics across different model pairs: MATH benchmark performance scores, percentage of large model usage, and the average number of parameters used during inference compared to using only the large model. Figure \ref{fig:config_grid}
 presents these results across four model pairs (LLaMA 3.2 1B/3B, LLaMA 3.2 3B/11B, Qwen 2.5 0.5B/14B, and Qwen 2.5 1.5B/14B), showing how performance varies with different entropy thresholds controlling model switching.

The results reveal a striking pattern across all model pairs: significant performance can be maintained while dramatically reducing the average number of parameters used during inference. In the LLaMA experiments, the 3B/11B pairing maintains 50.4\% MATH score (96.7\% of maximum performance) while using the larger model for only 43\% of tokens, reducing average parameters used during inference by 41.5\%. Similarly, the 1B/3B pairing achieves 45.2\% accuracy (97.4\% of maximum) with 45\% large model usage, reducing average parameters used by 36.5\% (Table \ref{tab:model_comparisons}).

These efficiency gains become particularly pronounced with larger size differentials in the Qwen family. The 1.5B/14B pairing achieves 74.3\% accuracy (92.9\% of maximum performance) while using the larger model for just 25\% of tokens, reducing average parameters used by 67\%. Most dramatically, the 0.5B/14B pairing reaches 69\% accuracy (86.3\% of maximum) with 35.3\% large model usage, reducing average parameters used by 62.4\%.

The data demonstrates a consistent non-linear relationship between large model usage and performance across all pairs. We observe rapidly diminishing returns in performance improvement beyond approximately 40\% large model usage, suggesting that the benefits of increased model capacity are not uniformly necessary throughout the generation process. This pattern holds remarkably consistent despite the varying size ratios between model pairs, from the modest 3x difference in the LLaMA 1B/3B pairing to the dramatic 28x difference in the Qwen 0.5B/14B pairing.

Further examination of entropy threshold patterns reveals several significant findings. Across all model pairs, we observe a consistent optimal range of $\tau=0.125$-$0.25$ that maximizes the trade-off between performance and parameter efficiency. This consistency across architectures and parameter scales suggests the existence of a fundamental boundary in computational complexity requirements for language generation tasks that transcends specific model implementations.

The efficiency gains of our methodology demonstrate super-linear scaling with respect to the parameter differential between model pairs. In transitioning from the relatively modest parameter ratio of the LLaMA 3B/1B pairing (3x) to the substantial differential of the Qwen 14B/0.5B pairing (28x), we observe that parameter reduction benefits nearly double (from 36.5\% to 62.4\% reduction) while maintaining comparable relative performance (97.4\% vs 86.3\% of maximum). This relationship suggests that EAD's effectiveness increases with parameter scale disparity, offering particular utility for efficient deployment of large-scale models.

\section{Discussion}
The effectiveness of Entropy Adaptive Decoding (EAD) can be understood through the empirical patterns observed in language model behavior during complex reasoning tasks. Our results demonstrate that computational requirements exhibit significant heterogeneity across the generation process, particularly in mathematical reasoning where cognitive demands are non-uniformly distributed.
Analysis of model behavior during mathematical problem-solving reveals distinct operational modes characterized by varying entropy levels. High-entropy regions strongly correlate with the initiation of new reasoning steps, where models must make crucial logical decisions. For instance, selecting an appropriate mathematical technique or initiating a new proof step consistently generates elevated entropy values, indicating increased prediction uncertainty. Conversely, subsequent elaboration and explanation phases show markedly lower entropy, suggesting these regions can be effectively handled by reduced model capacity.

This bimodal distribution of computational requirements explains EAD's ability to maintain performance while reducing large model usage. By dynamically allocating computational resources based on local entropy measurements, the system effectively matches model capacity to generation complexity. Critical reasoning junctures receive full model capacity, while routine exposition is handled by more efficient smaller models, enabling significant computational savings without proportional performance degradation.

Our findings identify a fundamental inefficiency in standard language model architectures: the allocation of uniform computational resources regardless of token-specific complexity. Traditional approaches apply identical computational depth to all tokens, whether generating complex mathematical insights or routine conjunctions. This uniform treatment results in substantial computational waste, particularly evident in mathematical reasoning tasks where near-peak performance is maintained while using the larger model for less than half of the tokens.

The magnitude of this inefficiency scales directly with model size - as models grow larger, the computational overhead for non-reasoning intensive tokens increases proportionally. This observation suggests current approaches to model scaling may be fundamentally inefficient, particularly for tasks with heterogeneous computational requirements. EAD addresses this through dynamic resource allocation, potentially indicating a more efficient path for model scaling.

While entropy proves effective as a computational complexity proxy, we acknowledge its limitations as a simplified metric of prediction uncertainty. It may fail to capture nuanced aspects of reasoning complexity or occasionally misidentify computational requirements. However, empirical results demonstrate that these theoretical limitations are outweighed by its practical utility in guiding model switching decisions. Future research directions could explore more sophisticated complexity metrics, such as attention pattern analysis or dedicated assessment circuits, though such approaches must balance increased precision against computational overhead.

These results have significant implications for language model architecture and deployment strategies. They suggest that current approaches to model inference may be unnecessarily conservative in their pursuit of uniform computational depth. The demonstration that heterogeneous computational allocation can maintain performance while reducing resource usage indicates potential directions for more efficient model scaling and deployment methodologies.

\section{Conclusion}
This work introduces Entropy Adaptive Decoding (EAD), a novel approach that challenges fundamental assumptions about computational resource allocation in language model inference. By dynamically switching between models based on prediction uncertainty, we demonstrate that current uniform-computation approaches significantly overallocate resources during generation.
Our extensive evaluation on the MATH benchmark reveals consistent patterns across model families and size differentials. The approach proves particularly effective with larger model pairs - using the LLaMA family, we maintain 96.7\% of the 11B model's performance while reducing its usage to 43\% of tokens. These gains amplify with greater size differentials in the Qwen family, where we preserve 92.5\% of the 14B model's capabilities while deploying it for only 25\% of tokens, reducing average parameter count by 67\%.

These findings suggest two key insights for language model deployment. First, they demonstrate that perfect output fidelity, while valuable, may come at an unnecessarily high computational cost for many applications. Second, they indicate that effective model scaling might benefit from focusing on strategic resource deployment rather than uniform computation increases. As language models continue to grow in size and computational demands, EAD offers a practical pathway toward making these capabilities more accessible in resource-constrained environments while maintaining most of their performance benefits.

Looking forward, this work opens new research directions in adaptive computation for language models. Future work might explore more sophisticated switching mechanisms, investigate optimal model pairings for specific tasks, and examine how these insights could inform model architecture design.


\bibliographystyle{unsrt}

\begin{thebibliography}{1}

\bibitem{noauthor_openai_nodate}
OpenAi.
\newblock {OpenAI} o1 {Hub} {\textbar} {OpenAI}.

\bibitem{shannon_prediction_1951}
C.~E. Shannon.
\newblock Prediction and {Entropy} of {Printed} {English}.
\newblock {\em Bell System Technical Journal}, 30(1):50--64, 1951.
\newblock \_eprint: https://onlinelibrary.wiley.com/doi/pdf/10.1002/j.1538-7305.1951.tb01366.x.

\bibitem{holtzman_curious_2020}
Ari Holtzman, Jan Buys, Li~Du, Maxwell Forbes, and Yejin Choi.
\newblock The {Curious} {Case} of {Neural} {Text} {Degeneration}, February 2020.
\newblock arXiv:1904.09751 [cs].

\bibitem{meister_locally_2023}
Clara Meister, Tiago Pimentel, Gian Wiher, and Ryan Cotterell.
\newblock Locally {Typical} {Sampling}, February 2023.
\newblock arXiv:2202.00666 [cs].

\bibitem{leviathan_fast_2023}
Yaniv Leviathan, Matan Kalman, and Yossi Matias.
\newblock Fast {Inference} from {Transformers} via {Speculative} {Decoding}, May 2023.
\newblock arXiv:2211.17192 [cs].

\bibitem{cai_medusa_2024}
Tianle Cai, Yuhong Li, Zhengyang Geng, Hongwu Peng, Jason~D. Lee, Deming Chen, and Tri Dao.
\newblock Medusa: {Simple} {LLM} {Inference} {Acceleration} {Framework} with {Multiple} {Decoding} {Heads}, June 2024.
\newblock arXiv:2401.10774 [cs].

\bibitem{chen_ee-llm_2024}
Yanxi Chen, Xuchen Pan, Yaliang Li, Bolin Ding, and Jingren Zhou.
\newblock {EE}-{LLM}: {Large}-{Scale} {Training} and {Inference} of {Early}-{Exit} {Large} {Language} {Models} with {3D} {Parallelism}, June 2024.
\newblock arXiv:2312.04916 [cs].

\bibitem{simonds_modem_2024}
Toby Simonds, Kemal Kurniawan, and Jey~Han Lau.
\newblock {MoDEM}: {Mixture} of {Domain} {Expert} {Models}, October 2024.
\newblock arXiv:2410.07490 [cs].

\bibitem{hendrycks_measuring_2021}
Dan Hendrycks, Collin Burns, Saurav Kadavath, Akul Arora, Steven Basart, Eric Tang, Dawn Song, and Jacob Steinhardt.
\newblock Measuring {Mathematical} {Problem} {Solving} {With} the {MATH} {Dataset}, November 2021.
\newblock arXiv:2103.03874 [cs].

\end{thebibliography}

\end{document}